\documentclass{article}
\usepackage{spconf,epsfig}
\usepackage{kotex}
\usepackage{cite}
\usepackage{amsfonts}
\usepackage{amssymb}
\usepackage{amsthm}
\usepackage[cmex10]{amsmath}
\usepackage{mathtools}
\usepackage{algpseudocode}
\usepackage{enumitem}
\usepackage{graphicx}
\usepackage{subcaption}
\usepackage{verbatim}
\usepackage{caption}
\usepackage{bm}
\usepackage{dblfloatfix}
\usepackage{booktabs}
\usepackage{paralist}
\usepackage{float}
\usepackage[ruled,vlined]{algorithm2e}
\usepackage{enumitem}
\usepackage{diagbox}

\let\OLDthebibliography\thebibliography
\renewcommand\thebibliography[1]{
  \OLDthebibliography{#1}
  \setlength{\parskip}{0pt}
  \setlength{\itemsep}{0pt plus 0.3ex}
}

\begin{document}\sloppy

\def\x{{\mathbf x}}
\def\L{{\cal L}}

\title{Forget-SVGD: Particle-Based Bayesian Federated Unlearning}
%
\name{Jinu Gong, Osvaldo Simeone, Rahif Kassab, and Joonhyuk Kang \vspace{-1cm}}
\address{}

\maketitle

\begin{abstract}
\vspace{-0.2cm}
Variational \emph{particle-based} Bayesian learning methods have the advantage of not being limited by the bias affecting more conventional parametric techniques. This paper proposes to leverage the flexibility of non-parametric Bayesian approximate inference to develop a novel Bayesian federated unlearning method, referred to as \emph{Forget-Stein Variational Gradient Descent (Forget-SVGD)}. Forget-SVGD builds on SVGD -- a particle-based approximate Bayesian inference scheme using gradient-based deterministic updates -- and on its distributed (federated) extension known as Distributed SVGD (DSVGD). Upon the completion of federated learning, as one or more participating agents request for their data to be ``forgotten'', Forget-SVGD carries out local SVGD updates at the agents whose data need to be ``unlearned'', which are interleaved with communication rounds with a parameter server. The proposed method is validated via performance comparisons with non-parametric schemes that train from scratch by excluding data to be forgotten, as well as with existing parametric Bayesian unlearning methods.
\end{abstract}
\begin{keywords}
Bayesian learning, Machine unlearning, Federated learning, Stein variational gradient descent.
\end{keywords}
\setlength{\abovedisplayskip}{3pt}
\setlength{\belowdisplayskip}{4pt}
\vspace{-0.1cm}
\section{Introduction}
\vspace{-0.2cm}

Federated learning trains shared models based on local data at participating agents through rounds of local computations and communications with a central server (see Fig. 1). While an agent may be initially willing to participate in a federated learning protocol, at a later stage it may decide to retract its authorization for its data to be implicitly encoded in -- and hence possibly retrieved from -- the shared model at the central server. The right to make such a decision is enshrined in ``right-to-be-forgotten'', or ``right-to-erasure'', laws such as the European Union's General Data Protection Regulation (GDPR)  \cite{ginart2019making}. An exact execution of the ``unlearning'' of data from an agent would require training the model from scratch by excluding the agents whose data are to be ``forgotten''. However, exact unlearning is inefficient in terms of communication cost and time, and \emph{machine unlearning} has been introduced to describe schemes that approximately remove the contribution of a subset of the overall data set via local optimization steps  \cite{cao2015towards,guo2019certified,ginart2019making,nguyen2020variational,golatkar2020eternal,sekhari2021remember,fu2021bayesian,thudi2021necessity}.

\begin{figure}[t]
\centering
\includegraphics[width=0.7
    \columnwidth]{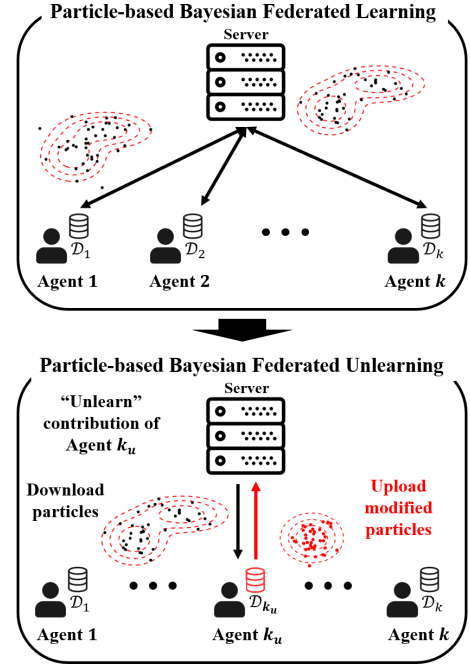}
\vspace{-0.1cm}
\caption{Machine learning and unlearning in a particle-based Bayesian federated learning framework.}
\label{fig:MG}
\vspace{-0.4cm}
\end{figure}


Federated learning protocols are conventionally designed within a frequentist framework, in which the goal of learning is optimizing over a shared model parameter vector. Frequentist learning is known to be limited in its capacity to quantify epistemic uncertainty, as well as support online and active learning (see, e.g., \cite{khan2021machine} and references therein). In contrast, Bayesian learning optimizes over distributions in the model parameter space, retaining information about epistemic uncertainty. As such, it is particularly well suited for federated learning applications, in which individual agents have limited data, causing significant levels of epistemic uncertainty, which needs to be quantified in order to ensure reliability and trustworthiness \cite{ruhe2019bayesian}. Federated Bayesian learning protocols have been proposed that apply variational parametric \cite{bui2018partitioned,vono2021qlsd} and non-parametric \cite{kassab2020federated} Bayesian schemes. Bayesian unlearning  was  studied in \cite{nguyen2020variational,gong2021bayesian,jose2021unified} for centralized settings by introducing variational unlearning criteria. These criteria were leveraged in \cite{gong2021bayesian} to develop a \emph{parametric} unlearning protocol for device-to-device networks.


In this paper, we consider a federated learning setting as illustrated in Fig. 1, and assume that the system implements any federated Bayesian learning protocol prior to the unlearning phase. Our goal is to develop a Bayesian \emph{unlearning} protocol that, starting from some shared trained model, is able to ``remove'' from it the contribution of a subset of agents. In order to leverage the flexibility of non-parametric methods to describe arbitrary distributions \cite{liu2016stein}, we adopt a \emph{particle-based} approach in which the variational distribution at the server is described by a number of particles. Note that this representation collapses to a standard frequentist formulation when a single particle is used, and that it is compatible with parametric federated learning protocols (for which particles can be produced via sampling).  

The novel Bayesian federated unlearning method, referred to as \emph{Forget-Stein Variational Gradient Descent (Forget-SVGD)}. Forget-SVGD builds on SVGD \cite{liu2016stein} -- a particle-based approximate Bayesian inference scheme using gradient-based deterministic updates -- and on its distributed (federated) extension known as Distributed SVGD (DSVGD) \cite{kassab2020federated}. Upon the completion of federated learning, as one or more participating agents request for their data to be ``forgotten'', Forget-SVGD carries out local SVGD updates at the agents whose data need to be ``unlearned'' interleaved with communication rounds with a parameter server. 



The rest of the paper is organized as follows. Sec. \ref{problem} introduces the setup and defines the variational Bayesian unlearning problem. Sec. \ref{background} and \ref{unlearning} review DSVGD and introduce the proposed Forget-SVGD protocol, respectively. Sec. \ref{experiments} describes numerical results, and Sec. \ref{conclusion} concludes the paper.

\vspace{-0.1cm}

\section{Problem Definition}
\vspace{-0.3cm}
\label{problem}
\subsection{Setup}
\vspace{-0.2cm}

As illustrated in Fig. \ref{fig:MG}, we consider a federated learning set-up with a set $\mathcal{K}=\{1,\ldots,K\}$ of $K$ agents communicating with a central node in a parameter server architecture. The local data set $\mathcal{D}_{k}=\{z_{k,n} \}_{n=1}^{N_k}$ of agent $k\in\mathcal{K}$ contains $N_k$ data points, and the associated training loss for model parameter $\theta$ is defined as
\begin{align}
L_k(\theta)=\frac{1}{N_k}\sum_{n=1}^{N_k} \ell_k(z_{k,n}|\theta)
\end{align}
for some loss function $\ell_k(z|\theta)$. We also denote as $\mathcal{D}=\bigcup_{k=1}^K\mathcal{D}_k$ the global data set.

In variational Bayesian federated learning, the agents collectively aim at obtaining a \textit{variational distribution} $q(\theta)$ on the model parameter space that minimizes the \textit{global free energy} (see, e.g., \cite{kassab2020federated,bui2018partitioned})
\begin{align}
\min_{q(\theta)} \bigg\{ F(q(\theta))\!=\!\sum_{k=1 }^K&\mathbb{E}_{\theta\sim q(\theta)}[L_k(\theta)]+\alpha\cdot \mathbb{D}\big( q(\theta)\big\|p_0 (\theta)\big) \bigg\},\vspace{-0.3cm}\label{eq:global_fe}
\end{align}
where $\alpha > 0$ is a ``temperature parameter"; $\mathbb{D}\left(\cdot\|\cdot\right)$ denotes Kullback–Leibler (KL) divergence; and $p_0(\theta)$ is a prior distribution. Minimizing the global free energy $F(q(\theta))$ seeks for a distribution $q(\theta)$ that is close to the prior $p_0 (\theta)$ while also minimizing the average sum-training loss. The unconstrained optimal solution of problem (\ref{eq:global_fe}) is given by the \textit{global generalized posterior distribution}
\begin{align}
q^*(\theta|\mathcal{D})&=\frac{1}{Z}\cdot \tilde{q}^*(\theta|\mathcal{D})\label{eq:opt_sol_ori}\\ \textrm{where}\quad\tilde{q}^*(\theta|\mathcal{D})&=p_0 (\theta) \exp \left(-\frac{1}{\alpha}\sum_{k=1}^K L_k(\theta)\right),\label{eq:opt_sol}
\end{align}
which coincides with the conventional posterior $p\big(\theta|\mathcal{D}\big)$ when we set $\alpha=1$ and the loss function is given by the log-loss $\ell_k(z|\theta)=-\log p(z|\theta)$.

In practice, problem (\ref{eq:global_fe}) can only be solved approximately by either: (\textit{i}) assuming a parametric form for the variational posterior $q(\theta|\mathcal{D})$, e.g., a Gaussian probability density function \cite{bui2018partitioned}; or (\textit{ii}) representing the variational posterior $q(\theta|\mathcal{D})$ in a non-parametric fashion based on a number of samples $\{\theta_n\}^{N}_{n=1}$ \cite{liu2016stein}.

\vspace{-0.3cm}
\subsection{Federated Machine Unlearning}
\vspace{-0.2cm}
In this paper, we focus on the problem of machine unlearning \cite{nguyen2020variational,sekhari2021remember}. Accordingly, the goal is to ``remove" information about the data of a subset $\mathcal{U}\subset \mathcal{K}$ of agents from the approximate solution $q(\theta|\mathcal{D})$ of the federated learning problem (\ref{eq:global_fe}). Clearly, one could obtain a variational posterior $q(\theta|\mathcal{D}_{-\mathcal{U}})$ by retraining of scratch using the data set $\mathcal{D}_{-\mathcal{U}}=\mathcal{D}\setminus \mathcal{D}_{\mathcal{U}}$ that excludes the data sets $\mathcal{D}_{\mathcal{U}}=\{\mathcal{D}_k \}_{k\in\mathcal{U}}$ from the agents whose data are to be ``forgotten". However, this may be costly in terms of computation and convergence time. Machine unlearning is concerned with developing more efficient unlearning protocols that do not require complete retraining from scratch.

We follow the variational unlearning formulation introduced in \cite{nguyen2020variational}, whereby unlearning of a data set $\mathcal{D}_{\mathcal{U}}$ from the variational posterior $q(\theta|\mathcal{D})$ is formulated as the minimization of the \textit{unlearning free energy}
\begin{align}
\min_{q(\theta)} \bigg\{ F_{\mathcal{U}}(q(\theta))= \sum_{k\in\mathcal{U}}&\mathbb{E}_{\theta\sim q(\theta)}[-L_k (\theta)]\nonumber\\
&+\alpha\cdot \mathbb{D}(q(\theta)\|q(\theta|\mathcal{D}))\bigg\}.\label{eq:ref_unlearning}
\end{align}
Note that, unlike the global free energy $F(q(\theta))$ in (\ref{eq:global_fe}), the unlearning free energy $F_{\mathcal{U}}(q(\theta))$ in (\ref{eq:ref_unlearning}) includes as its first term the negative of the training loss of the data to be removed, while the role of the prior $p_0 (\theta)$ in (\ref{eq:global_fe}) is played in (\ref{eq:ref_unlearning}) by the variational posterior $q(\theta|\mathcal{D})$. Intuitively, minimization of the unlearning free energy aims at finding a distribution $q(\theta)$ that is close to the current variational posterior $q(\theta|\mathcal{D})$, while maximizing the average training loss for the data sets to be forgotten. 

In previous work \cite{nguyen2020variational,gong2021bayesian}, parametric methods were studied for learning and unlearning, while in this paper we focus on non-parametric technique owing to their enhanced flexibility and reduced bias.
\vspace{-0.1cm}

\section{Particle-Based Federated Learning}
\label{background}
\vspace{-0.2cm}

In this section, we review non-parametric variational federated learning as introduced in \cite{kassab2020federated}. Specifically, reference \cite{kassab2020federated} presents DSVGD, which represents the variational posterior $q(\theta|\mathcal{D})$ via $N$ particles, $\{\theta_n\}_{n=1}^N$, and extends SVGD \cite{liu2016stein} to a federated learning setup within the framework of partitioned variational inference (PVI) \cite{bui2018partitioned}.

\vspace{-0.3cm}
\subsection{Distributed SVGD}
\vspace{-0.2cm}

In DSVGD, the server maintains a set of $N$ particles $\{\theta_n \}_{n=1}^N$ that are iteratively updated by a subset of agents. We focus here on the case of a single agent scheduled at each iteration, since the extension to more than one agent is direct as detailed in \cite{kassab2020federated}. DSVGD minimizes the global free energy (\ref{eq:global_fe}) over the variational distribution $q(\theta)$ in a distributed manner. At the beginning of the $i$-th iteration, the server stores the current particles $\{\theta_n^{(i-1)} \}_{n=1}^N$, which represent the current iterate $q^{(i-1)}(\theta)$ of the global variational distribution. An explicit estimate of distribution $q^{(i-1)} (\theta)$ can be obtained, e.g., via kernel density estimator (KDE) with some kernel function $K(\theta, \theta')$. Following expectation propagation (EP) \cite{gelman2014expectation} and PVI, DSVGD writes the variational distribution $q^{(i-1)}(\theta)$ is interpreted as being factorized as $q^{(i-1)}(\theta)=p_0(\theta)\prod_{k=1}^K t_k^{(i-1)}(\theta)$, where the term $t_k^{(i-1)} (\theta)$ is known as approximate likelihood of agent $k$. At each iteration $i$, the scheduled agent $k$ updates the variational distribution $q^{(i-1)}(\theta)$ by modifying its approximate likelihood to a new iterate $t_{k}^{(i)}(\theta)$.

To this end, at each iteration $i$, the scheduled agent $k$ updates the current set of particles $\{\theta_n^{(i-1)} \}_{n=1}^N$ with the goal of minimizing the \textit{local free energy}
\begin{align}
 \min_{q(\theta)} \bigg\{F_k^{(i)}(q(\theta))=\mathbb{E}_{\theta\sim q(\theta)}[L_{k}(\theta)]+\alpha \mathbb{D}\big(q(\theta)\big\| \hat{p}_k^{(i)}(\theta)\big)\bigg\},\label{eq:FVU_min}
\end{align}
where the so-called cavity distribution $\hat{p}_k^{(i)}(\theta)$ \cite{kassab2020federated}
\begin{align}
\hat{p}_k^{(i)}(\theta)\propto\frac{q^{(i-1)}(\theta)}{t_{k}^{(i-1)}(\theta)}\label{eq:cavity}
\end{align}
removes the approximate local likelihood of agent $k$, which is updated as 
\begin{align}
t_k^{(i)}(\theta)=\frac{q^{(i)}(\theta)}{q_{k}^{(i-1)}(\theta)} t_{k}^{(i-1)}(\theta).\label{eq:llh}
\end{align}
In DSVGD, the approximate likelihood is also represented by a set of particles, which is local to agent $k$. The detailed steps are as follows.

\textbf{Initialization.} Draw the set of $N$ global particles $\{\theta_n^{(0)} \}_{n=1}^N$ from the prior $p_0(\theta)$; and initialize at random the set of local particles for all agents $k\in\mathcal{K}$.

\textbf{Step 1.} Server schedules an agent $k\in\mathcal{K}$. Agent $k$ downloads the current global particles $\{\theta_n^{(i-1)} \}_{n=1}^N$ from the server.

\textbf{Step 2.} Agent $k$ initialize $\{\theta_{n}^{[0]}=\theta_{n}^{(i-1)}\}_{n=1}^N$ and applies SVGD to address the problem of minimizing (\ref{eq:FVU_min}) over the set of global particles as
\begin{align}
\theta_n^{[l]}&\leftarrow\theta_{n}^{[l-1]}+\epsilon \phi\left({\theta_n^{[l-1]}}\right),\label{eq:unlearn_svgd_particle}
\end{align}
for $n=1,\ldots, N$ where 
\begin{align}
\phi(\theta)=\frac{1}{N}\sum_{j=1}^N \bigg[\kappa\left(\theta_j^{[l-1]},\theta\right)&\nabla_{\theta_j}\log \tilde{p}_k^{(i)}\left(\theta_j^{[l-1]}\right)\nonumber\\
&+\nabla_{\theta_j}\kappa\left(\theta_j^{[l-1]},\theta\right)\bigg],\label{eq:phi_hat}
\end{align}
for local iteration $l=1,\ldots,L$ and a positive definite kernel $\kappa (\cdot, \cdot)$. The gradient of the tilted distribution $\tilde{p}_k^{(i)}(\theta)\propto\hat{p}_{k}^{(i)}(\theta) \exp \left(-\alpha^{-1} L_k(\theta)\right)$ can be computed by using the KDE of $q^{(i-1)}(\theta)$ and $t_k^{(i-1)}(\theta)$, i.e., $q^{(i-1)}(\theta)= \frac{1}{N} \sum_{n=1}^N K(\theta, \theta_n^{(i-1)})$ and $t_k^{(i-1)}(\theta)= \frac{1}{N} \sum_{n=1}^N K(\theta, \theta_{k,n}^{(i-1)})$.

\textbf{Step 3.} Agent $k$ sets $\{\theta_n^{(i)}=\theta_n^{[L]}\}_{n=1}^N$. The updated global particles $\{\theta_n^{(i)} \}_{n=1}^N$ are sent to the server that sets $\{\theta_n =\theta_n^{(i)} \}_{n=1}^N$. Finally, agent $k$ updates its local particles $\{\theta_{k, n}^{(i)} \}_{n=1}^N$ via $L_{\textrm{local}}$ ``distillation" steps as
\begin{align}
\theta_{k,n}^{[l]}&\leftarrow\theta_{k,n}^{[l-1]}+\epsilon_{k} \phi_{k}\left({\theta_{k,n}^{[l-1]}}\right),\label{eq:update_local_particle}
\end{align}
where $\epsilon_{k}$ is learning rate for local particle update, and
\begin{align}
\hat{\phi}_{k}^{*}(\theta)=\frac{1}{N}\sum_{j=1}^N \bigg[\kappa&\left( \theta_{k,j}^{[l-1]},\theta \right)\nabla_{\theta_{k,j}}\log t_k^{(i)}\left(\theta_{k,j}^{[l-1]}\right)\nonumber\\
&\qquad\qquad+\nabla_{\theta_{k,j}}\kappa\left(\theta_{k,j}^{[l-1]},\theta\right)\bigg],\label{eq:phi_hat_local}
\end{align}
for local iteration $l=1,\ldots,L_{local}$, with the gradient $\nabla_{\theta}\log t_k^{(i)}\left(\theta\right)=\nabla_{\theta}\log q_k^{(i)}\left(\theta\right)-\nabla_{\theta}\log q_k^{(i-1)}\left(\theta\right)+\nabla_{\theta}\log t_k^{(i-1)}\left(\theta\right)$ evaluated using the particles $\{\theta_n^{(i)} \}_{n=1}^N$, $\{\theta_n^{(i-1)} \}_{n=1}^N$ and $\{\theta_{k,n}^{(i-1)} \}_{n=1}^N$. Agent $k$ updates sets $\{\theta_{k,n}^{(i)}=\theta_{k,n}^{[L]}\}_{n=1}^N$, while the other agents $k'\neq k$ set $\{\theta_{k',n}^{(i)} =\theta_{k',n}^{(i-1)} \}_{n=1}^N$.
\vspace{-0.1cm}

\section{Particle-Based Machine Unlearning: Forget-SVGD}
\vspace{-0.2cm}
\label{unlearning}

In this section, we introduce a particle-based machine unlearning algorithm based on SVGD, which we refer to as Forget-SVGD. Forget-SVGD addresses problem (\ref{eq:ref_unlearning}) by representing the variational posterior $q(\theta|\mathcal{D}\setminus \mathcal{D}_{\mathcal{U}})$ through a set of global particles $\{\theta_n \}_{n=1}^N$. The initial values of these particles before unlearning may be obtained via DSVGD, as described in the previous section; or by parametric Bayesian federated learning methods, such as PVI \cite{bui2018partitioned} followed by sampling from the obtained parametric distribution $q(\theta|\mathcal{D})$ (see Sec. \ref{experiments}).

In order to ``forget" the data of agents in set $\mathcal{U}$, Forget-SVGD schedules an agent $k\in\mathcal{U}$ at each iteration $i$. A scheduled agent $k$ downloads the current set of $N$ particles $\{\theta_n^{(i-1)} \}_{n=1}^N$ from the server, and updates these to minimize the \textit{local unlearning free energy}, i.e.,
\begin{align}
 \min_{q(\theta)}\bigg\{ \tilde{F}_k^{(i)}(q(\theta))\!=\!\mathbb{E}_{\theta\sim q(\theta)}&[-L_{k}(\theta)]+\alpha \mathbb{D}\big(q(\theta)\big\| \hat{p}_k^{(i)}(\theta)\big)\bigg\},\label{eq:FVU_min_unlearn}
\end{align}
where the cavity distribution $\hat{p}_k^{(i)}(\theta)$ is defined in (\ref{eq:cavity}). Forget-SVGD operates as follows.

\textbf{Initialization.} The initial set of $N$ particles $\{\theta_n^{(0)}\}_{n=1}^N$ represents the variational distribution obtained as a result of Bayesian federated learning; initialize at random local particles $\{\theta_{k,n}^{(0)} \}_{n=1}^N$ for all agents $k\in \mathcal{U}$

\textbf{Step 1.} At iteration $i$, the server schedules an agent $k\in\mathcal{U}$ in the set of agents whose data must be unlearned. Agent $k$ downloads the current global particles $\{\theta_n^{(i-1)} \}_{n=1}^N$ from the server.

\textbf{Step 2.} Agent $k$ initialize $\{\theta_{n}^{[0]}=\theta_{n}^{(i-1)}\}_{n=1}^N$ and updates downloaded particles as (\ref{eq:unlearn_svgd_particle})-(\ref{eq:phi_hat}) with the caveat that the (unnormalized) tilted distribution $\tilde{p}_k^{(i)}(\theta)$ is defined as
\begin{align}
\tilde{p}_k^{(i)}(\theta)=\frac{q^{(i-1)}(\theta)}{t_k^{(i-1)}(\theta)} \exp \left(\frac{1}{\alpha} L_k(\theta)\right)\label{eq:tilted_dist},
\end{align}
where $q^{(i-1)}(\theta)$ and $t_k^{(i-1)}(\theta)$ are computed by using the respective KDEs with global and local particles, respectively.

\textbf{Step 3.} Agent $k$ sets $\{\theta_n^{(i)}=\theta_n^{[L]}\}_{n=1}^N$. The updated particles $\{\theta_n^{(i)} \}_{n=1}^N$ are sent to the server that sets $\{\theta_n = \theta_n^{(i)} \}_{n=1}^N$. Agent $k$ updates its local particles $\{\theta_{k, n}^{(i)} \}_{n=1}^N$ via $L_{\textrm{local}}$ distillation steps as (\ref{eq:update_local_particle})-(\ref{eq:phi_hat_local}). Finally, agent $k$ updates the current local particles as $\{\theta_{k,n}^{(i)}=\theta_{k,n}^{[L]}\}_{n=1}^N$, while the other agents $k'\neq k$ set $\{\theta_{k',n}^{(i)} =\theta_{k',n}^{(i-1)} \}_{n=1}^N$.

\vspace{-0.1cm}

\section{Experiments}
\label{experiments}
\vspace{-0.2cm}

In this section, we evaluate the performance of Forget-SVGD using synthetic and real-world examples. For all experiments, as in \cite{liu2016stein}, we assume the radial basis function (RBF) kernel $\kappa (x,x_0)=\textrm{exp}(-\| x-x_0 \|_{2}^2 /h)$ and the bandwidth $h=\textrm{med}^2/\log N$, where $\textrm{med}$ is the median of the pairwise distances between the particles in the current iteration. We also use the Gaussian kernel for the KDE with a bandwidth $\lambda=0.55$. We fix the temperature parameter $\alpha=1$, and use AdaGrad as in \cite{liu2016stein}.

\vspace{-0.3cm}
\subsection{Benchmark}
\vspace{-0.2cm}
 For comparison, we consider parametric variational methods for both learning and unlearning, namely PVI \cite{bui2018partitioned} and unlearning PVI (UL-PVI) \cite{gong2021bayesian}, respectively. In PVI, the variational distribution $q(\theta)$ is parametrized as $p(\theta|\eta)=\textrm{ExpFam}(\theta|\eta)$, where $\eta$ represents the natural parameter vector. At each iteration $i$, the server, which maintains the current iterate $\eta^{(i-1)}$, schedules an agent $k$. The scheduled agent $k$ downloads global natural parameter iterate $\eta^{(i-1)}$, and updates it through local iterations $l=1,...,L$ with initialization $\eta^{[0]}=\eta^{(i-1)}$ via the natural gradient descent rule on the local free energy (\ref{eq:FVU_min}), yielding the update \cite{bui2018partitioned}
\begin{align}
{\eta^{[l]}}\!\leftarrow\!{\eta^{[l-1]}}-&\epsilon\left( {\eta_{k}^{[l-1]}}+\frac{1}{\alpha}\!\cdot\!\nabla_{\mu^{[l-1]}}\mathbb{E}_{ q(\theta|\eta^{[l-1]})}[L_{k}(\theta)]\right),\label{eq:non_conj_update}
\end{align}
where $\eta^{[L]}=\eta^{(i)}$, $\epsilon$ and $\mu^{[l-1]}$ are learning rate and the moment parameter corresponding to the natural parameter $\eta^{[l-1]}$, respectively. Then, it updates local natural parameter as
\begin{align}
\eta_{k}^{(i)}=\eta^{(i)}-\eta^{(i-1)}+\eta_{k}^{(i-1)},\label{eq:loc_nat_update}
\end{align}
and uploads updated global natural parameter $\eta^{(i)}$ to the server; while the other agents $k'\neq k$ set the local natural parameter $\eta_{k'}^{(i)}=\eta_{k'}^{(i-1)}$.

Starting from the obtained global natural parameter $\eta^{*}$, in UL-PVI, at iteration $i$, the server schedules an agent $k$ in set $\mathcal{U}$, and the agent $k$ updates global natural parameter via natural gradient descent applied to the unlearning free energy (\ref{eq:ref_unlearning}) as
\begin{align}
{\eta^{[l]}}\!\leftarrow\!{\eta^{[l-1]}}-&\epsilon\left( {\eta_{k}^{[l-1]}}+\frac{1}{\alpha}\cdot\nabla_{\mu^{[l-1]}}\mathbb{E}_{ q(\theta|\eta^{[l-1]})}[-L_{k}(\theta)]\right),\label{eq:non_conj_unlearn}
\end{align}
for $l=1,\ldots,L$ with $\eta^{[0]}=\eta^{(i-1)}$ and $\eta^{[L]}=\eta^{(i)}$. Then, it updates local natural parameter $\eta_{k}^{(i)}$ using (\ref{eq:loc_nat_update}), and uploads updated global natural parameter $\eta^{(i)}$ to the server, while the other agents $k'\neq k$ set the local natural parameter $\eta_{k'}^{(i)}=\eta_{k'}^{(i-1)}$.

\vspace{-0.3cm}
\subsection{Mixture of Gaussians}
\vspace{-0.2cm}

\begin{figure*}[t]
\centering
\vspace{-0.3cm}
\includegraphics[width=\textwidth]{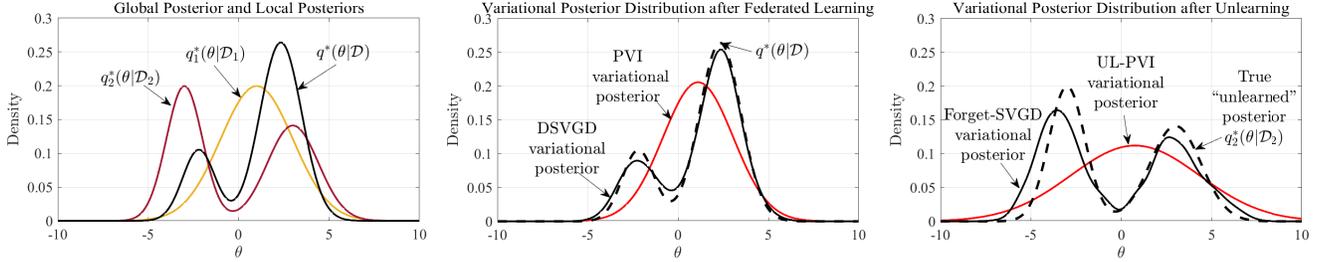}
\vspace{-0.5cm}
\caption{True posterior $q^*(\theta|\mathcal{D})$ and local posteriors $q_1^*(\theta|\mathcal{D}_1)$ and  $q_2^*(\theta|\mathcal{D}_2)$ (left); Variational global posterior distribution after federated learning using PVI or DSVGD and true global posterior $q^*(\theta|\mathcal{D})$ (middle); Variational posterior distribution after unlearning the data of agent $1$ using UL-PVI or Forget-SVGD, along with the true ``unlearned" posterior $q_2^*(\theta|\mathcal{D}_2)$ (right).}
\label{fig:three graphs}
\vspace{-0.4cm}
\end{figure*}

To start, we consider $K=2$ agents for which the local posteriors $q_k^* (\theta|\mathcal{D}_k)\propto p_0(\theta)\exp(-\frac{1}{\alpha}L_k (\theta))$ for $k=1,2$ are defined as $q_1^*(\theta|\mathcal{D}_1)\propto p_0 (\theta)\mathcal{N} (\theta|1,4)$ and $q_2^*(\theta|\mathcal{D}_2)\propto p_0 (\theta)\left(\mathcal{N} (\theta|-3,1)+\mathcal{N} (\theta|3,2)\right)$, where the prior $p_0 (\theta)$ is uniform over $[-10, 10]$, i.e., $p_0(\theta)=\mathcal{U} (\theta|-10,10)$. In the federated learning phase, we update the approximate global posterior using DSVGD with $N=500$ particles \cite{kassab2020federated}, and then apply Forget-SVGD for unlearning of the data of agent 1. For PVI/UL-PVI, we choose the variational posteriors as Gaussian distributions with prior $p_0 (\theta)=\mathcal{N}(0, 16)$. The top part of Fig. \ref{fig:three graphs} shows the true global posterior $q^*(\theta|\mathcal{D})$ in (\ref{eq:opt_sol_ori})-(\ref{eq:opt_sol}) and local posteriors $q_1^*(\theta|\mathcal{D}_1)$, $q_2^*(\theta|\mathcal{D}_2)$; the middle part illustrates the optimized variational posterior distribution after learning using PVI (red line) and DSVGD (black line), along with the corresponding true global posterior $q^*(\theta|\mathcal{D})$ (black dashed line); and the bottom part depicts the optimized variational posterior distribution after unlearning the data of agent $1$ using UL-PVI (red line) and Forget-SVGD (black line), along with the corresponding ideal ``unlearned" posterior $q^*(\theta|\mathcal{D}_2)=q_2^*(\theta|\mathcal{D}_2)$ (black dashed line). DSVGD approximates the true posterior better than PVI due to the inherent bias of PVI caused by its parametric assumption on the variational posterior. This advantage is inherited by Forget-SVGD, which provides a closer approximation of the true unlearning posterior than UL-PVI.

\vspace{-0.3cm}
\subsection{Multi-Label Classification}
\vspace{-0.2cm}

\begin{figure}[t]
\centering
\includegraphics[width=0.9
    \columnwidth]{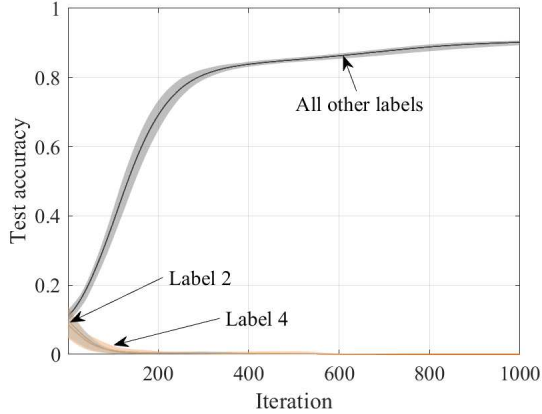}
\vspace{-0.2cm}
\caption{Accuracy on the test dataset for label 2 and label 4, which are to be unlearned, and on the remaining labels for exact unlearning based on training from scratch using the data set $\mathcal{D}\setminus \mathcal{D}_u$. The shaded area covers the support of the accuracy in 10 experiments.}
\label{fig:acc_retrain}
\vspace{-0.4cm}
\end{figure}
We now consider multi-label classification with a Bayesian Neural Networks (BNN) models on the MNIST dataset. We adopt a ``non-iid" setting with $K=5$ agents by assigning each agent $100$ examples from only two of the ten classes of MNIST images. Each agent observes data from two different labels so as all labels are observed across the $k=5$ agents. The model consists of one fully-connected hidden layer of 100 hidden neurons and a softmax layer. Following \cite{kristiadi2020being}, we perform Bayesian learning only on the last layer while the other parameters are fixed. To this end, we first pre-train the network $\theta_{\textrm{MAP}}$ using the dataset $\mathcal{D}=\bigcup_{k=1}^K\mathcal{D}_k$, and then we apply Bayesian learning only to the last layer. For unlearning, the dataset $\mathcal{D}_u$ of agent $u$ having labels 2 and 4 is to be ``forgotten".

\begin{figure}[t]
\centering
\includegraphics[width=0.9
    \columnwidth]{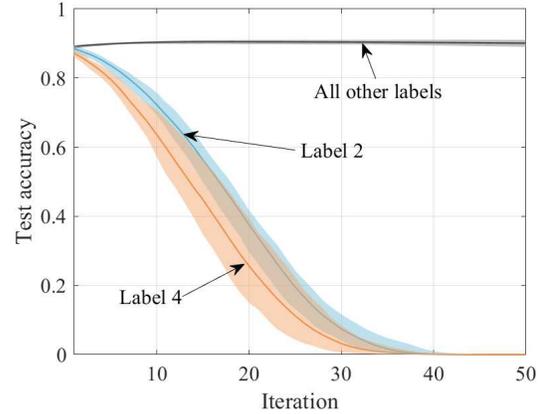}
\vspace{-0.2cm}
\caption{Accuracy on the test dataset for label 2 and label 4, which are to be unlearned, and on the remaining labels during unlearning via Forget-SVGD. The shaded area covers the support of the accuracy in 10 experiments.}
\label{fig:acc_unlearn}
\vspace{-0.4cm}
\end{figure}

Before considering the performance of Forget-SVGD, as a benchmark, Fig. \ref{fig:acc_retrain} shows the test accuracy under exact unlearning based on training from scratch by using the data set $\mathcal{D}\setminus \mathcal{D}_u$. For training from scratch, we apply SVGD on the last layer by using remaining dataset, while the other parameters are fixed by pre-trained values. The blue and orange lines depict the test accuracy for test data corresponding to the labels (label 2 and label 4) to be unlearned, while the black line is the average test accuracy for all the other labels. The shaded area depicts the maximum and minimum accuracy among 10 experiments. In Fig. \ref{fig:acc_unlearn}, we present the same metrics for Forget-SVGD. Forget-SVGD is shown to be able to unlearn $\mathcal{D}_u$ in around $40$ iterations in the sense that the test accuracy of the unlearning labels vanishes, while the test accuracy for the remaining labels is maintained. This result is not only $25$ times faster than that of training from scratch, but also, unlike training from scratch, it only requires the participation of the two agents whose data need to be unlearned.

\vspace{-0.1cm}

\section{conclusions}
\label{conclusion}
\vspace{-0.2cm}
This paper has introduced a particle-based variational Bayesian unlearning algorithm -- Forget-SVGD -- that  inherits from particle-based Bayesian learning techniques the flexibility in representing arbitrary distributions. Forget-SVGD was shown via experiments to significantly improve the communication and time efficiency of exact unlearning via training from scratch.

\bibliographystyle{IEEEbib}
\bibliography{refs}

\begin{thebibliography}{10}

\bibitem{ginart2019making}
Antonio Ginart, Melody~Y Guan, Gregory Valiant, and James Zou,
\newblock ``{Making AI forget you: Data deletion in machine learning},''
\newblock {\em arXiv preprint arXiv:1907.05012}, 2019.

\bibitem{cao2015towards}
Yinzhi Cao and Junfeng Yang,
\newblock ``{Towards making systems forget with machine unlearning},''
\newblock in {\em 2015 IEEE Symposium on Security and Privacy}, 2015, pp.
  463--480.

\bibitem{guo2019certified}
Chuan Guo, Tom Goldstein, Awni Hannun, and Laurens Van Der~Maaten,
\newblock ``{Certified data removal from machine learning models},''
\newblock {\em arXiv preprint arXiv:1911.03030}, 2019.

\bibitem{nguyen2020variational}
Quoc~Phong Nguyen, Bryan Kian~Hsiang Low, and Patrick Jaillet,
\newblock ``{Variational Bayesian unlearning},''
\newblock {\em arXiv preprint arXiv:2010.12883}, 2020.

\bibitem{golatkar2020eternal}
Aditya Golatkar, Alessandro Achille, and Stefano Soatto,
\newblock ``{Eternal sunshine of the spotless net: Selective forgetting in deep
  networks},''
\newblock in {\em Proceedings of the IEEE/CVF Conference on Computer Vision and
  Pattern Recognition}, 2020, pp. 9304--9312.

\bibitem{sekhari2021remember}
Ayush Sekhari, Jayadev Acharya, Gautam Kamath, and Ananda~Theertha Suresh,
\newblock ``{Remember what you want to forget: Algorithms for machine
  unlearning},''
\newblock {\em arXiv preprint arXiv:2103.03279}, 2021.

\bibitem{fu2021bayesian}
Shaopeng Fu, Fengxiang He, Yue Xu, and Dacheng Tao,
\newblock ``Bayesian inference forgetting,''
\newblock {\em arXiv preprint arXiv:2101.06417}, 2021.

\bibitem{thudi2021necessity}
Anvith Thudi, Hengrui Jia, Ilia Shumailov, and Nicolas Papernot,
\newblock ``{On the Necessity of Auditable Algorithmic Definitions for Machine
  Unlearning},''
\newblock {\em arXiv preprint arXiv:2110.11891}, 2021.

\bibitem{khan2021machine}
Mohammad~Emtiyaz Khan,
\newblock ``{Machine learning from a Bayesian perspective},''
\newblock {\em https://emtiyaz.github.io/papers/MLfromBayes.pdf}, 2021.

\bibitem{ruhe2019bayesian}
David Ruhe, Giovanni Cina, Michele Tonutti, Daan de~Bruin, and Paul Elbers,
\newblock ``{Bayesian modelling in practice: Using uncertainty to improve
  trustworthiness in medical applications},''
\newblock {\em arXiv preprint arXiv:1906.08619}, 2019.

\bibitem{bui2018partitioned}
Thang~D Bui, Cuong~V Nguyen, Siddharth Swaroop, and Richard~E Turner,
\newblock ``{Partitioned variational inference: A unified framework
  encompassing federated and continual learning},''
\newblock {\em arXiv preprint arXiv:1811.11206}, 2018.

\bibitem{vono2021qlsd}
Maxime Vono, Vincent Plassier, Alain Durmus, Aymeric Dieuleveut, and Eric
  Moulines,
\newblock ``{QLSD: Quantised Langevin stochastic dynamics for Bayesian
  federated learning},''
\newblock {\em arXiv preprint arXiv:2106.00797}, 2021.

\bibitem{kassab2020federated}
Rahif Kassab and Osvaldo Simeone,
\newblock ``{Federated generalized Bayesian learning via distributed Stein
  variational gradient descent},''
\newblock {\em arXiv preprint arXiv:2009.06419}, 2020.

\bibitem{gong2021bayesian}
Jinu Gong, Osvaldo Simeone, and Joonhyuk Kang,
\newblock ``{Bayesian variational federated learning and unlearning in
  decentralized networks},''
\newblock in {\em 2021 IEEE 22nd International Workshop on Signal Processing
  Advances in Wireless Communications (SPAWC)}, 2021, pp. 216--220.

\bibitem{jose2021unified}
Sharu~Theresa Jose and Osvaldo Simeone,
\newblock ``{A unified PAC-Bayesian framework for machine unlearning via
  information risk minimization},''
\newblock in {\em Proc. IEEE Int. Workshop on Machine Learning for Signal
  Processing (MLSP)}, 2021.

\bibitem{liu2016stein}
Qiang Liu and Dilin Wang,
\newblock ``{Stein variational gradient descent: A general purpose Bayesian
  inference algorithm},''
\newblock in {\em Advances in Neural Information Processing Systems}. 2016,
  vol.~29, Curran Associates, Inc.

\bibitem{gelman2014expectation}
Andrew Gelman, Aki Vehtari, Pasi Jyl{\"a}nki, Christian Robert, Nicolas Chopin,
  and John~P Cunningham,
\newblock ``{Expectation propagation as a way of life},''
\newblock {\em arXiv preprint arXiv:1412.4869}, vol. 157, 2014.

\bibitem{kristiadi2020being}
Agustinus Kristiadi, Matthias Hein, and Philipp Hennig,
\newblock ``{Being bayesian, even just a bit, fixes overconfidence in relu
  networks},''
\newblock in {\em International Conference on Machine Learning}. PMLR, 2020,
  pp. 5436--5446.

\end{thebibliography}

\end{document}